# Bi-directional Shape Correspondences (BSC): A Novel Technique for 2-d Shape Warping in Quadratic Time?

AbdulRahman Oladipupo Ibraheem

rahmanoladi@yahoo.com

Computing and Intelligent Systems Research Group, Obafemi Awolowo University, Ile-Ife, Nigeria.

**Abstract**

We propose Bidirectional Shape Correspondence (BSC) as a possible improvement on the famous shape contexts (SC) framework. Our proposals derive from the observation that the original SC framework enforces a one-to-one correspondence between sample points, and that this leads to two possible drawbacks. First, this denies the framework of the opportunity to effect a many-to-many matching between points on the two shapes being compared. Second, this calls for the Hungarian algorithm which unfortunately usurps $O(N^3)$ time. While the dynamic-space-warping dynamic programming algorithm has provided a standard solution to the first problem above, dynamic programming demands $O(N^5)$ for general multi-contour shapes and $O(wN^2)$ for the special case of single-contour shapes, even after an heuristic search window of width $w$ has been chosen. Therefore, in this work, we propose a simple method for computing "many- to –many" correspondences, for the class of all 2-d shapes, in quadratic time. Our approach is to explicitly let each point on the first shape choose a best match on the second shape, and vice versa. Along the way, we also propose the use of data-clustering techniques for dealing with the outliers problem. From another viewpoint, it turns out this clustering can be viewed as an autonomous, rather than pre-computed, sampling of shape boundary.

## 1    Introduction

Shape contexts was introduced by Belongie *et. al.* in [1]. Shape contexts is applicable to two fundamental problems of 2-d vision: computation of correspondences and shape matching. Although shape contexts, when combined with the thin plate spline interpolator, continues to boast one of the best results (0.6% error rate) on the MNIST database of handwritten digits [2], there appears to be room for improvement in its overall framework, especially given that it does not perform as well on certain other databases. The shape contexts framework seeks to enforce a one-to-one correspondence between sample points on the two shapes being matched. As illustrated in Figures 1a and 1b, this leads to problems in certain situations, as follows. In Figure 1a, $p_1, p_2, \ldots, p_6$ represent a subset of equally-spaced sample points on the boundary of the shape in the figure. Likewise, in Figure 1b, $q_1, q_2, \ldots, q_6$ represent a subset of equally-spaced sample points on the boundary of the shape in the figure. When asked, an human expert is more likely to choose the correspondences:
$p_1 \leftrightarrow q_1, p_2 \leftrightarrow q_2, p_3 \leftrightarrow q_2, p_4 \leftrightarrow q_3, p_5 \leftrightarrow q_4, p_5 \leftrightarrow q_5, p_6 \leftrightarrow q_6$, between the two shapes. There are two key points to note here. First, the human expert maps two different points, $p_2$ and $p_3$, on the first shape to a single point $q_2$ on the second shape. Second, the human expert maps two different points, $q_4$ and $q_5$, on the second shape to a single point $p_5$ on the first shape. Clearly, the choice of the human expert goes contrary to the approach in [1] which sought to enforce one-to-one correspondences. Moreover, in an attempt to achieve one-to-one correspondences, the authors of [1] were naturally led to employ the Hungarian algorithm which unfortunately usurps $O(N^3)$ time, where $N$ is the number of sample points on each shape.

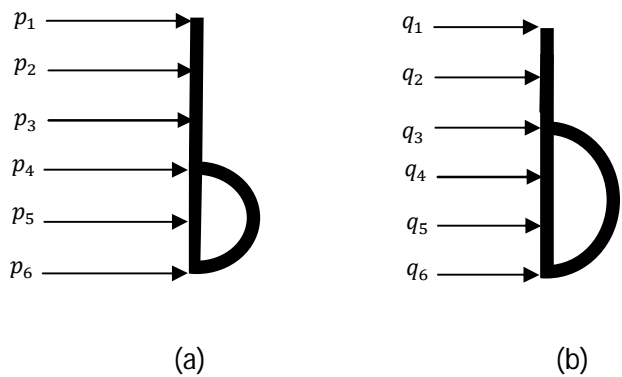

(a)          (b)

Fig. 1 : How the enforcement of one-to-one correspondences between two shapes can go contrary to the choice of an human expert. For the two shapes above, an human expert is more likely to choose the ***non*** one-to-one correspondences: $p_1 \leftrightarrow q_1, p_2 \leftrightarrow q_2, p_3 \leftrightarrow q_2, p_4 \leftrightarrow q_3, p_5 \leftrightarrow q_4, p_5 \leftrightarrow q_5, p_6 \leftrightarrow q_6$ than a one-to-one correspondence, such as $p_1 \leftrightarrow q_1, p_2 \leftrightarrow q_2, p_3 \leftrightarrow q_3, p_4 \leftrightarrow q_4, p_5 \leftrightarrow q_5, p_6 \leftrightarrow q_6$.

For the restricted case of single contour shapes, Adamek and O'Connor [2] have responded to the above highlighted problem, via their dynamic- space- warping dynamic programming algorithm, which directly allows for many-to-many correspondences. However, dynamic programming requires $O(N^5)$ time for general multi-contour shapes, and $O(wN^2)$ time for single contour shapes, even after an heuristic search window of width $w$ has been chosen.

In this work, we will present a simple approach to computing 2-d shape correspondences in $O(M^2)$ time, with $M = \max\{m, n\}$, where $m$ and $n$ are respectively the number of boundary pixels on the first and second shapes. As shall be seen, our approach leads to many-to-many correspondences for both single contour and multi-contour shapes. In addition, we propose the use of a data-clustering method, such as the Otsu method [3], for pruning correspondences, by grouping correspondences into "good" and "bad" matches. Furthermore, we do not pre-sample boundary points, rather, we leave it to our framework to autonomously select an optimal sampling in the course of its action. The hope is that these contributions might lead to improved performance on benchmark databases such as the MNIST database of handwritten images and the MPEG data base of silhouettes. Finally, although we shall present our approach in the framework of shape contexts and the thin plate spline interpolator, we believe it is general enough to accommodate other frameworks such as those of [2], [4] and [5].

## 2      Formal details of our proposed method

Let $P$ and $Q$ be the contours of two arbitrary shapes in the plane. We begin by arbitrarily taking the "direction" from $P$ to $Q$ as the forward direction, and the direction from $Q$ to $P$ as the backward direction. To move on, let $P$ contain $m$ points: $P = \{p_1, p_2, \ldots, p_m\}$; and let $Q$ contain $n$ points: $Q = \{q_1, q_2, \ldots, q_n\}$. For each $p_i \in P$, we define the **forward correspondence point** for $p_i$, amongst the points in $Q$, as the point $q_{k_i} \in Q$ satisfying:

$$q_{k_i} = \mathop{argmin}_{q_j \in Q} C(p_i, q_j) \qquad 1$$

In the above equation, $C(p_i, q_j)$ is the cost of matching $p_i$ and $q_j$ using shape contexts (other approaches such as those in [2], [4] and [5] are also possible); the reader is referred to [1] for details of computing $C(p_i, q_j)$ using shape contexts. Ideally, the costs $C(p_i, q_j)$, $i = 1, 2, \ldots, m$, $j = 1, 2, \ldots, n$ should be pre-computed and stored in an $m$ by $n$ matrix, $M(i, j)$. Naturally, the entry at the $i$-th row and $j$-th column of $M(i, j)$ stores the cost $C(p_i, q_j)$. Computing matrix $M(i, j)$ requires $O(m^2 + n^2) = O(M^2)$ time, where $M = \max\{m, n\}$. Now, in Equation 1 above, $p_i$ is associated with the $i$-th row, $R_i$, of $M(i, j)$, and it should not be hard to see that $q_{k_i}$ is associated with the $k$-th column, which holds the least score in $R_i$. Thus computing $q_{k_i}$ distills to a simple search for the location of a/the minimum value in a row vector $R_i$. Clearly, this search takes $O(n)$ time. We will need to execute this search for each $p_i \in P$. Thus, we need a total of $O(mn) = O(M^2)$ time. At this juncture, we may define the **average forward match cost**, $\vec{C}$, from $P$ to $Q$ as:

$$\vec{C} = \frac{1}{m}\sum_{i=1}^{m} C(p_i, q_{k_i}) \qquad 2$$

We view correspondences computed by Equation 1 above as **forward correspondences** from $P$ to $Q$, because we allowed each point on $P$ to choose a best match on $Q$. We can also compute **backward correspondences** from $Q$ to $P$ by letting each $q_j \in Q$ choose a best match on $P$. For each $q_j \in Q$, we define the **backward correspondence point** for $q_j$, amongst the points in $P$, as the point $p_{k_j} \in P$ satisfying $\hat{P}$:

$$p_{k_j} = \mathop{argmin}_{p_i \in P} C(p_i, q_j) \qquad 3$$

The required searches for the $p_{k_j}$'s proceed along column vectors in matrix $M(i,j)$, in contrast to how the searches for the $q_{k_i}$'s proceed along row vectors. Nonetheless, the searches for the $q_{k_i}$'s require $O(M^2)$ time, in consonance with that of the $p_{k_j}$'s. Furthermore, we define the **average backward match cost**, $\overleftarrow{C}$, from $Q$ to $P$ as:

$$\overleftarrow{C} = \frac{1}{n}\sum_{j=1}^{n} C(p_{k_j}, q_j) \qquad 4$$

There are two key sub-problems to address: computation of correspondences and computation of an overall match score between the two shapes in question. Taking a top-down approach, we address the latter before the former. Specifically, we simply define the **bi-directional** match cost, or bidirectional shape correspondence, $\overline{C}$, between $P$ and $Q$ as the average of $\vec{C}$ and $\overleftarrow{C}$.

$$\overline{C} = \frac{1}{2}(\vec{C} + \overleftarrow{C}) \qquad 5$$

We turn now to the correspondence problem. A problem with correspondences is that of outliers. To address this problem, we propose that the match costs be clustered into two groups, a high match group and a low match group, using a data clustering technique, such as the Otsu Method. We explain this procedure using the forward direction. First, observe that the forward costs, $C(p_i, q_{k_i})$, $i = 1,2,\ldots,m$, can be viewed as an $m$-element array, from which we can choose the best correspondence point, $q_{k_i}$, for each $p_i \in P$. The idea is to cluster this $m$-element costs array into two groups, with the low cost group representing the good matches and the high cost group representing the erroneous bad matches. After the clustering, we obtain a set of costs, $\widehat{C}(p_i, q_{k_i})$, $i = g_1, g_2, \ldots, g_u$, $u \leq m$, for the good-match group. Notice that the above clustering also indirectly clusters the underlying set $P$ into two groups, one associated with the good matches and the other with the bad matches. We denote the subset of $P$ associated with the good matches by $\hat{P} = \{p_{g_1}, p_{g_2}, \ldots, p_u\}$. Next, with respect to $\widehat{C}(p_i, q_{k_i})$, we define a function $F: \hat{P} \to Q$, such that, for each $p_i \in \hat{P}$, we have $F(p_i) = q_{k_i}$. With this, we can now compute the **pruned** forward correspondence point for each point $p_i \in \hat{P}$ as the point $q_{k_i} \in Q$ satisfying:

$$q_{k_i} = F(p_i) \qquad 6$$

The preceding permits the definition of a **pruned average forward match cost**, $\widehat{\vec{C}}$, from $P$ to $Q$ as follows:

$$\widehat{\vec{C}} = \frac{1}{u}\sum_{i=1}^{u} \widehat{C}(p_i, q_{k_i}) \qquad 7$$

By analogy, we have backward counterparts for Equations 6 and 7. Specifically, in the backward direction, after clustering, we obtain good-match costs, $\widehat{C}(p_{k_i}, q_j)$, a set $\widehat{Q}$, a function $G: \widehat{Q} \to P$, such that, for each $q_i \in \widehat{Q}$, we have $G(q_i) = p_{k_i}$, and, finally, a definition for the **pruned average backward match cost**, $\widehat{\widehat{C}}$, from $Q$ to $P$ :

$$\widehat{\widehat{C}} = \frac{1}{v}\sum_{j=1}^{v} \widehat{C}(p_{k_j}, q_j) \qquad 8$$

To proceed, we must decide whether to choose the forward correspondences or the backward correspondences. A natural way to make this decision is to choose the forward correspondences if $\widehat{\widehat{C}} \leq \widehat{\widehat{C}}$, but to choose the backward correspondences otherwise. Now, we recall that our aim is to achieve dynamic space warping (i.e. many-to-many mapping) in quadratic time. Without loss of generality, using the forward correspondences, we shall now explain how our framework can achieve dynamic space warping. The function $F: \widehat{P} \to Q$ underpins the computation of correspondences in the forward direction. Clearly, this function allows a many-to-one mapping from $\widehat{P}$ to $Q$, so it remains to discuss how a one-to-many mapping can go from $\widehat{P}$ to $Q$ in the forward direction. Actually, because $F$ is a mathematical function, we cannot not get a one-to-many mapping with $F$ in the strict sense. However, our framework permits something that closely approximates a one-to-many mapping. We exhibit this in Figure 2.

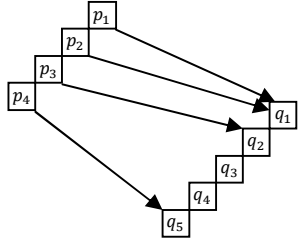

Fig. 2 : Illustration of how our framework tends to achieve many-to-many correspondences between (sections of ) the boundaries of two shapes. The $p_i$'s and $q_i$'s are pixels on the boundaries of the first and second shapes respectively. The correspondences $p_1 \leftrightarrow q_1$ and $p_2 \leftrightarrow q_1$ form a many-to-one correspondence. On the other hand, $p_3 \leftrightarrow q_2$ and $p_4 \leftrightarrow q_5$ can be interpreted as warping the boundary segment between $p_3$ and $p_4$ to the boundary segment between $q_2$ and $q_5$, thus somehow indicating the one-to-many correspondences: $p_3 \leftrightarrow q_2, p_3 \leftrightarrow q_3, p_4 \leftrightarrow q_4, p_4 \leftrightarrow q_5$.

Our aim for computing the correspondences is to use them for calculating a thin plate spline (TPS) interpolator between the shapes underlying $P$ and $Q$. We refer the reader to [1] for core information on how the TPS is used for this purpose. Quite comparable to the Lagrangian interpolator which can be used to fit a polynomial to set of points in the two-dimensional plane, TPS can be used to fit a surface to a set of points in three-dimensional space. For our purposes herein, TPS can either be used in the forward direction, or in the backward, depending on whether $\widehat{\widehat{C}} \leq \widehat{\widehat{C}}$ or not. To fix the idea, let us consider the forward direction. For that direction, we need to sample $N$ points, $p_{t_1}, p_{t_2}, \ldots, p_{t_N}$ from the set $\widehat{P}$. We denote by $x_i$ the $x$-co-ordinate of the point $p_{t_i}$, and by $y_i$ the $y$-co-ordinate of $p_{t_i}$. Further, we denote by $x'_i$ the $x$-co-ordinate of $q_{k_i}$, and by $y'_i$ its $y$-co-ordinate. Then, the inputs to the TPS procedure are the constraints $f_x(x_i, y_i) = x'_i$ and $f_y(x_i, y_i) = y'_i$, $i = t_1, t_2, \ldots, t_N$. The outputs from the procedure are the functions $f_x: R^2 \to R$ and $f_y: R^2 \to R$. With this, the shape underlying $P$ can be warped unto the shape underlying $Q$. After this warping, we would have completed an iteration of shape contexts –TPS pair of computations. Similar to what is done in [1], we propose that this iteration be done about three times, plus a final shape contexts computation.

We now pause to emphasize the key aspects of our contributions. First, as earlier explained, our framework allows dynamic space warping, unlike the original work of [1] who favored a one-to-one correspondence. Second, unlike the approach in [3, 4, 5], which leads to an $O(N^5)$ algorithm (where $N$ is number of sampled boundary points on each shape) for multi-contour shapes, our approach takes quadratic $O(M^2)$ time, where $M = \max\{m, n\}$. Furthermore, we highlight two subtle points about our approach. Firstly, during each shape contexts–TPS iteration, the bi-directional match cost between the two shapes does not take into consideration the effect of the clustering process. This is a deliberate arrangement aimed at penalizing false matches. On the contrary, input to the TPS interpolator involves only those points that are considered good-matches by the clustering process. The reason why we do this is to improve the accuracy of the TPS warping between the two shapes. Secondly, notice that towards the computation of correspondences, we do not sample boundary points; the only reason why we sample points is towards the computation of the TPS interpolator. However, the effect of the clustering operation on the boundary points can be viewed as an autonomous sampling of the boundary. While boundary point sampling improves computational speed, it nevertheless does not improve the order of growth of the algorithm.